\pdfoutput=1

\documentclass[twocolumn]{article}
\usepackage{arxiv}
\usepackage[font=small]{caption}
\usepackage{enumitem}
\usepackage{cancel}

\usepackage[
backend=biber,
style=alphabetic,
sorting=nty,  
block=space, 
]{biblatex}
\AtEveryBibitem{
 \clearfield{file}
 \clearfield{eprint}
 \clearfield{keywords}
 \clearfield{abstract}
 \clearfield{day}
 \clearfield{urlyear}
 
 \ifentrytype{book}{}{
  \clearlist{publisher}
  \clearname{editor}
 }
}

\usepackage{amsmath,amssymb}
\DeclareMathOperator{\E}{\mathbb{E}}
\DeclareMathOperator{\bigH}{\mathrm{H}}
\DeclareMathOperator{\DKL}{\mathbf{D}_{KL}}
\newcommand{\Eq}[1]{Eq.~\eqref{#1}}
\newcommand{\Fig}[1]{Fig.~\ref{#1}}

\newcommand{\qx}{q(x)}

\newcommand{\po}{p(o)}
\newcommand{\pref}{\tilde{p}}
\newcommand{\bfF}{\mathbf{F}}
\newcommand{\calG}{\mathcal{G}}
\newcommand{\calH}{\mathcal{H}}
\newcommand{\calL}{\mathcal{L}}

\usepackage{url}            
\usepackage{booktabs}       
\usepackage{nicefrac}       
\usepackage{aligned-overset}
\usepackage{graphicx}
\usepackage{array} 
\usepackage{microtype}

\usepackage{hyperref}
\hypersetup{
    linkcolor=blue,
    citebordercolor=0.3 0.9 0.7,
    urlbordercolor=0.2 0.6 0.7,
    linkbordercolor=0.6 0.1 0.1,
}

\title{Free Energy Risk Metrics for Systemically Safe AI: Gatekeeping Multi-Agent Study}

\author{
 Michael Walters \\
  Gaia Lab \\
  Nuremberg, Germany
 \And
 Rafael Kaufmann \\
  Primordia Co.\\
  Cascais, Portugal \\
 \And
 Justice Sefas \\
  University of British Columbia \\
  B.C., Canada \\ 
 \And
 Thomas Kopinski \\
  Gaia Lab \\
  Fachhochschule Südwestfalen \\
  Meschede, Germany\\
}

\date{November 2024}

\begin{document}

\maketitle

\begin{abstract}
    We investigate the Free Energy Principle as a foundation for measuring risk in agentic and multi-agent systems. From these principles we introduce a Cumulative Risk Exposure metric that is flexible to differing contexts and needs. We contrast this to other popular theories for safe AI that hinge on massive amounts of data or describing arbitrarily complex world models. In our framework, stakeholders need only specify their preferences over system outcomes, providing straightforward and transparent decision rules for risk governance and mitigation. This framework naturally accounts for uncertainty in both world model and preference model, allowing for decision-making that is epistemically and axiologically humble, parsimonious, and future-proof. We demonstrate this novel approach in a simplified autonomous vehicle environment with multi-agent vehicles whose driving policies are mediated by gatekeepers that evaluate, in an online fashion, the risk to the collective safety in their neighborhood, and intervene through each vehicle's policy when appropriate.
    We show that the introduction of gatekeepers in an AV fleet, even at low penetration, can generate significant positive externalities in terms of increased system safety.
\end{abstract}

\section{Introduction}

Rooted in physics, the Free Energy Principle (FEP), in tandem with Bayesian inference of world models, offers a compelling foundation in the Active Inference (ActInf) formulation of intelligent systems 
\cite{dacostaActiveInferenceDiscrete2020, fristonDesigningEcosystemsIntelligence2024, gottwaldTwoKindsFree2020, hyland2024freeenergy, kaufmannActiveInferenceModel2021, leibfriedVariationalInferenceModelFree2022a, millidgeWhenceExpectedFree2021, parrActiveInferenceFree2022}. 
One of the earliest progenitors of this idea is the \textit{Helmholtz machine}, proposed by Dayan, Hinton, Neal, and Zemel in 1995 \cite{dayanHelmholtzMachine1995}, connecting the statistical mechanics governing the Helmholtz Free Energy and perceptual processing. Here, treating the log-likelihood of perceptrons in a neural model as energy akin to statistical mechanics, learning proceeds as the minimization of variational free energy (VFE) through variational inference\footnote{
As the authors in \cite{gottwaldTwoKindsFree2020} point out, though VFE is not the same as Helmholtz Free Energy, the two concepts can be formally related.
}.
Fast-forward to the present day and the Bayesian Brain hypothesis has found popularity in neurosymbolic modeling, whereby perception and other decision/control mechanisms are driven by predictive (generative) models and hierarchical Bayesian uncertainty-resolving directives \cite{bayesianBrain2010, parrGeneralisedFreeEnergy2019}.
For an enriching summary around FEP and its connections to Bayesian/Active Inference, see Gottwald and Braun \cite{gottwaldTwoKindsFree2020}.

The FEP in Active Inference can be applied in a few different ways \cite{millidgeWhenceExpectedFree2021, gottwaldTwoKindsFree2020}, and interpreted in many more \cite{dacostaActiveInferenceDiscrete2020}.
These interpretations are variations on a classic theme: exploitation vs.\ exploration. Whether it's accuracy vs.\  complexity, risk vs.\ ambiguity, intrinsic value vs.\ extrinsic value, model evidence vs.\ information gain, or energy vs.\ entropy, the mechanics of the FEP live in the tension of this duality. 

To illustrate the rich connection between probabilistic modeling and the FEP, we begin with the common setup of an agent making observations $o_t$ at time $t$, and wishing to infer the latent state of the world $x_t$ through actions $a_t$ according to policy $\pi$ (which we will take as Markovian).
The agent's uncertainty about $x_t$ given its observations is expressed as the posterior \mbox{$p(x_t|o_t) = p(o_t,x_t)/p(o_t)$}.
With the standard assumption of the intractability of $p(o_t)$, Variational Inference prescribes we instead work with a tractable approximation, $q(x_t)$ that \textit{can} be computed.

Typically, the mismatch between $p(x)$ and $q(x)$ is quantified by the \textit{Kullback-Leibler divergence}, 
\begin{equation*}
    \DKL(q||p) = \int_{x} q(x)\ln\left(\frac{q(x)}{p(x)} \right) dx.
\end{equation*}

We will drop the subscript $t$ going forward in most cases when it is irrelevant. The KL divergence is convex for fixed $p$. Thus, the problem is recast with a new proxy objective: the minimization of $\DKL(q||p)$ through inference on $q$.

Finally, the KL divergence between the variational approximation of the true posterior $\DKL\bigl(q(x)||p(x|o)\bigr)$ has an intrinsic connection to the log-evidence $\ln p(o)$:
\begin{align}
    \DKL\bigl(q(x)||p(x|o)\bigr) &= \int_{x} \qx\ln\left(\frac{\qx \po}{p(x,o)} \right) dx \notag \\
    &= - \int_{x} q(x)\ln p(x,o)dx \notag \\
    &\hspace{2em} + \int_{x} \qx \ln\qx dx \notag \\
    &\hspace{2em} + \int_{x} \qx \ln\po dx \notag \\
    \vspace{1em}
    &\hspace{-6em} \Rightarrow \E_{\qx}\left[\ln\qx - \ln p(x,o) \right] + \ln \po. \label{eq:evidence-1}
\end{align}

In line (\ref{eq:evidence-1}) we make use of the fact that $\po$ is independent of $\qx$. Rearranging, we can express the evidence as
\begin{align*}
    \ln\po &= \DKL\bigl(q(x)||p(x|o)\bigr) - \E_{\qx}\bigl[\ln\qx \\
    &\hspace{2em} - \ln p(x,o) \bigr] \\
    &= \DKL\bigl(q(x)||p(x|o)\bigr) - \bfF(q).
\end{align*}

The $-\bfF (q)$ term gives a floor for the evidence (since \mbox{$\DKL(q||p) \geq 0$}), and as the evidence $\ln\po$ is fixed with respect to $\qx$, minimizing $\bfF(q)$ drives the floor up and \textit{minimizes} the KL divergence between $q$ and $p$.

As mentioned earlier, the free energy in statistical mechanics is, abstractly, the sum of an \textit{accuracy} term (energy), and a \textit{complexity} term (entropy). For example, for some distribution $\phi$, the Helmholtz Free Energy,
$$
F_H(\phi) = \langle E \rangle_\phi + \frac{1}{\beta}\bigH(\phi)
$$
where inverse temperature $\beta$ plays a weighting factor between energy and entropy. It is this similarity in form why $\bfF(q)$ is also called the \textit{variational free energy} (VFE):
\begin{align}
    \bfF(q) &= -\E_{\qx}[\ln p(x, o)] + \E_{\qx}[\ln\qx] \notag \\
    &= \underbrace{-\E_{\qx}[\ln p(x, o)]}_{\text{``Energy''}} - \bigH[\qx].
    \label{eq:vfe}
\end{align}

The entropic term is a form of Occam's razor, encouraging models to make fewer assumptions or have too many extraneous parameters. It also functions like a regularizer against overfitting to model evidence by the energy term. 
In the ActInf framework, agents are driven to reduce ``surprisal''---the discrepancy between their models and the world, i.e.\ VFE---primarily through two means (\cite{parrActiveInferenceFree2022} \S 2.6, \cite{millidgeWhenceExpectedFree2021}): 
\begin{itemize}[left=2em]
    \item (Perception) Updating world models to better fit the evidence.
    \item (Action) Exploration and actions in the world to elicit desirable outcomes or reduce uncertainty.
\end{itemize}

With a generative model $p(x, o)$, artificial agents can simulate potential futures and use the expected free energy to evaluate policies and inform their decisions.

\subsection{Extending into the future}

The VFE-based objective discussed thus far has focused on deriving a variational model $\qx$ through inference that both explains the data and is balanced by an entropic term.
However, this falls short of how a fully equipped ActInf agent would operate intelligently: using preference-biased predicted futures to inform its actions.
We defer the philosophical justification \cite{parrActiveInferenceFree2022}, but in sum, incorporating a preference prior distribution $\pref (o)$ over expected outcomes (or states $\pref (x)$) embeds the goal directives of the agent into the objective---elevating it from being just a Bayesian evidence-building machine.

Inference then proceeds towards minimizing the \textit{Expected Free Energy} (EFE) across candidate policies, where quality of fit is judged by the expected log likelihood of \textit{desired} observations, and exploration is encouraged through maximizing the divergence between the expected variational posterior and the expected variational prior
\footnotemark.
\footnotetext{We use an approximation here that the true and approximate posteriors are similar, $q(x|o) \approx p(x | o)$, meaning inference was successful. Without this assumption, there is an additional divergence term between these two quantities.}
\begin{align}
    \textbf{EFE}_t &\equiv \E_{q(o_t,x_t|\pi)}\left[\ln q(x_t|\pi) - \ln \pref (o_t, x_t)\right] \label{eq:EFE-def} \\ 
    &\approx - \underbrace{\E_{q(o_t, x_t|\pi)} \big[
    \ln\pref(o_t) \big]}_{\text{Extrinsic Value}} \label{eq:EFE-2} \\
    &\hspace{2em} - \underbrace{\E_{q(o_t|\pi)} \DKL \big[
    q(x_t | o_t) || q(x_t | \pi) \big]}_{\text{Epistemic Value}} \notag
\end{align}

where $\pref (o_t, x_t) = p(o_t | x_t)\pref (x_t)$. 
Taking a temporal mean-field factorization of the variational posterior \mbox{$q(x_{t:\tau}, \pi)\approx q(\pi)\prod_{s=t}^\tau q(x_s|\pi)$} and generative model \mbox{$\pref(o_{t:\tau}, x_{t:\tau})\approx \prod_{s=t}^\tau \pref(o_s)q(x_s|o_s)$}, severs the temporal dependence between steps, meaning the optimal path is that with the lowest sum $\sum_t \textbf{EFE}_t$.

Millidge, Tschantz, and Buckley \cite{millidgeWhenceExpectedFree2021} give considerable contemplation to the question of extending the VFE into the future and the natural origins of the EFE\footnotemark.
\footnotetext{
    Ultimately, it is argued that the EFE---the go-to formulation in ActInf---is by no means mandatory and even less natural of a construction than their proposed \textit{Free Energy of the Expected Future} (FEEF) alternative (not to be confused with the FEF discussed in this manuscript). 
    The difference being that the extrinsic value of the EFE is the maximization of the log model evidence (\Eq{eq:EFE-2}), whereas in the FEEF it is the minimization of the KL-divergence between the likelihood of observations predicted under a \textit{veridical generative model} and the marginal likelihood of observations predicted under the \textit{biased generative model}---which is argued to be more aligned with the goals of an ActInf agent.
} 
The authors go on to introduce an additional FEP-based formulation, the \textit{Free Energy of the Future} (FEF), which has an objective driven by the \textit{minimization} of the entropic term, in stark contrast to epistemic maximization: 
\begin{align}
    \textbf{FEF}_t &\equiv \E_{q(o_t,x_t|\pi)}\left[
        \ln q(x_t|o_t) - \ln \pref (o_t, x_t)
    \right] \label{eq:FEF-def} \\
    &\approx - \E_{q(o_t, x_t|\pi)} \big[
    \ln\pref(o_t | x_t) \big] \notag \\
    &\hspace{2em} + \E_{q(o_t|\pi)} \DKL \big[
    q(x_t | o_t) || q(x_t | \pi) \big]
    \label{eq:FEF-2}
\end{align}

Note the epistemic terms between the EFE and FEF differ only in their sign. Encouraging the minimization of an information-seeking term seems anathema to an ActInf agent, yet minimizing the FEF satisfies the FEP-driven goals of 1) bounding the model evidence (surprisal), and 2) minimizing the divergence between a variational posterior and a target model (whether that is based on the true world distribution or a preference prior in the context of Active Inference). 

\section{Cumulative Risk Exposure}
\label{sec:risk-define}

We propose and showcase an arrangement that repurposes and reframes the VFE construction laid out above. The canonical Active Inference agent begins with a known preference prior that informs its actions as expected VFE computations. However, by obfuscating the preference prior from the agent---or at least the \textit{true} stakeholder preference prior, if we still want the agent to operate in an ActInf fashion with its own preference prior---we can help buffer against certain reward specification pitfalls, like reward hacking, etc. 
In essence, this defines a Gatekeeper (GK) arrangement, where the GK has access to the agent's policies and can compute a policy's expected free energy according to its hidden preference prior as a form of policy evaluation and risk metric. 
Expressing values as preference prior distributions allows for a wide range of preference structures, including risk-aversion, social preferences, and non-Markovian utility functions \cite{skalseLimitationsMarkovianRewards2023}.

The free energy risk metric can be utilized as context prescribes, and we demonstrate a simple method whereby a risk threshold is defined as a point of criticality demanding gatekeeper intervention\footnotemark. To our knowledge, this is the first VFE-based gatekeeper model for agentic AI applications. 
\footnotetext{
A binary risk threshold is not the only option. Specifically, in a setting with continuous control variables, it would be possible to perform a smooth handover (linear combination) between agent policy and gatekeeper policy. This introduces complexities in the simulation model and will be left to future work.
}

When defining a risk metric, both the FEF and the EFE provide viable options. For contexts where exploration is \textit{discouraged}, the FEF offers a better form since its objective is minimized through low-entropy futures. This may be the better choice for safety-critical applications where minimizing unexpected behavior is preferred. Conversely, in domains with significant structural uncertainty and ambiguity (such as research and corporate strategy) or where downside risk is not deemed significant (such as arts and entertainment), an EFE formulation would encourage exploration.

\subsection{Adapting for observation-space}
Often it is the case that a preference prior is expressed in terms of outcomes, not hidden states. Thus, it useful to express the VFE formulae in observation-space. From the definition of EFE in \Eq{eq:EFE-def} (dropping the time-dependence),
\begin{align*}
    &\E_{q(o,x|\pi)}\bigl[
        \ln q(x|\pi) - \ln \pref (o, x)
    \bigr] \\
    &= \E_{q(o,x|\pi)}\bigl[
        \ln q(x|\pi) - \ln\pref(o) - \ln q(x|o)
    \bigr] \\
    &= \E_{q(o,x|\pi)}\bigl[
        \cancel{\ln q(x|\pi)} - \ln\pref(o) - \ln q(o|x) \\
        &\hspace{2em} - \cancel{\ln q(x|\pi)} + \ln q(o|\pi) 
    \bigr] \\
    &= \E_{q(o,x|\pi)}\bigl[
        - \ln\pref(o) - \ln q(o|x) + \ln q(o|\pi) 
    \bigr] \\
    &= -\underbrace{\E_{q(o,x|\pi)}\bigl[\ln \pref(o)\bigr]}_{\text{Extrinsic}} - \underbrace{\E_{q(x|\pi)}\bigl[\DKL[q(o|x) || q(o|\pi)\bigr]}_{\text{Information Gain}}
\end{align*}

making use of the definitions $\pref(x,o) = q(x|o)\pref(o)$ and $q(x,o|\pi) = q(x|\pi)q(o|x) = q(o|\pi)q(x|o)$, and Bayes' rule. Computationally, one can estimate these values through sampling of the variational prior and the produced observations.
Similar decompositions can be achieved for the FEF:
\begin{align*}
    &\E_{q(o,x|\pi)}\bigl[\ln q(x|o) - \ln \pref (o, x)\bigr] \\
    &=\E_{q(o,x|\pi)}\bigl[
        \ln q(x|o) - \ln\pref(o|x) - \ln q(x|\pi)
    \bigr] \\
    &=\E_{q(o,x|\pi)}\bigl[
        \ln q(o|x) + \cancel{\ln q(x|\pi)} - \ln q(o|\pi) \\
        &\hspace{2em} - \ln\pref(o|x) - \cancel{\ln q(x|\pi)}
    \bigr] \\
    &=\E_{q(o,x|\pi)}\bigl[
        \ln q(o|x) - \ln q(o|\pi) - \ln\pref(o|x)
    \bigr] \\
    &=\underbrace{-\E_{q(o,x|\pi)}\bigl[\ln\pref(o|x)\bigr]}_{\text{Extrinsic}}
    + \underbrace{E_{q(x|\pi)}\bigl[\DKL[q(o|x)||q(o|\pi)]\bigr]}_{\text{Information Gain}}
\end{align*}

Between the two decompositions, we see the sign flip on the epistemic term persist.

Finally, the VFE risk formulations thus far are lacking a balancing variable that weights the epistemic and extrinsic components. In analogy with free energy formulations of thermodynamics, we can introduce an inverse ``temperature'' to balance the terms of our risk equation. In abstract, the instantaneous risk at time $t$, for a variable set $\phi=[q, \pref, \pi]$ is
\begin{align}
    \calG_t(\phi) = \langle E \rangle_{\phi,t} \pm \frac{1}{\beta} \calH[\phi,t], \label{eq:risk-abs}
\end{align}

where $E$ and $\calH$ are the energetic and entropic components\footnotemark.
Recall, the EFE and FEF are expected free energy forms, which can be $\gamma$ time-discounted in aggregation across time. 
We thus define the \textit{Cumulative Risk Exposure} (CRE)
\begin{align}
    \calG_\Sigma(\phi, t) = \sum_{t'}^\tau \gamma^{t'} \calG_{t+t'}(\phi),
    \label{eq:risk-sum}
\end{align}

though we will commonly drop the time subscript in our discussions.

\footnotetext{The inverse temperature has an interesting parallel with the Probability Dependency Graph framework, where a $\beta$ term represents the degree of confidence/belief in a distribution \cite{richardsonLossInconsistencyProbabilistic2022}.
In our construction, confidence in $\pref$ can factored into $1/\beta$, but the inverse temperature carries a slightly different implication: one could be entirely confident in $\pref$ but still value including entropic contributions.}

\subsection{Preference prior construction}

\label{sec:pref-prior}
Choice of preference prior is context-dependent, but a natural form is the Boltzmann distribution
over some loss function $\calL$:
\begin{equation}
    \pref(\calL) = e^{-\beta\calL} / Z, \label{eq:pref}
\end{equation}
\begin{equation*}
    Z = \sum_j e^{-\beta\calL_j}.
\end{equation*}

With this formulation, the inverse temperature term in \Eq{eq:risk-abs}, which serves to balance the extrinsic and intrinsic terms, equivalently operates on the extrinsic, preference-based term instead of the intrinsic term, 
\begin{align*}
    \calG &= - \E_q\bigl[\ln \pref\bigr] \pm \frac{1}{\beta}\calH \\ 
    &\Rightarrow \beta \E_q\bigl[\calL \bigr] \pm \calH + \ln(Z).
\end{align*}

Consequently, from \Eq{eq:pref} $\beta$ quantifies a \textit{tolerance} to loss, scaling $\pref$ accordingly, and can be thought of as a \textit{preference temperature} of our system.
Very strong preference biases create a ``low temperature'' (high $\beta$) system that is very \textit{energetically sensitive} to preference alignment; conversely, weak preference bias creatures a smoothed out preference distribution that is more \textit{entropy dominated}, with lower energetic sensitivity.

Further, the Boltzmann distribution has the property that the ratio of state probabilities
\begin{equation*}
    \frac{\pref(\calL_1)}{\pref(\calL_2)} = \exp(-\beta({\calL_1 - \calL_2})).
\end{equation*}

Thus, we can calibrate $\beta$ from a maximum and minimum loss range, and those corresponding stakeholder-assigned desirabilities,
\begin{equation*}
    \ln \left(\frac{\pref_{max}}{\pref_{min}}\right) = -\beta(\calL_{max} - \calL_{min})
\end{equation*}
\begin{equation*}
    \Rightarrow \beta = \frac{\ln \left(\pref_{min} / \pref_{max}\right)}{\calL_{max} - \calL_{min}} \geq 0.
\end{equation*}

$\beta$ is non-negative since since by definition the desirability $\pref_{min} \geq \pref_{max}$, and $\calL_{max} \geq \calL_{min}$.

\subsection{Extending the approach}

As discussed by Hyland et al.\ \cite{hyland2024freeenergy}, minimizing a \textit{joint free energy} as a sum of individual agent free energies can avail game-theoretically optimized solutions that would otherwise not be played in selfish policies. Indeed, joint free energy minimization has been postulated as a potential core mechanism behind collective agency in biological systems \cite{shreeshaStressSharingCognitive2024, mcmillenCollectiveIntelligenceUnifying2024}. It is also translatable to the Cooperative Inverse Reinforcement Learning paradigm \cite{hadfield-menellCooperativeInverseReinforcement2024}, as agents model the preferences of humans and themselves.
In our AV experiment, the free energy of neighboring vehicle gatekeepers is aggregated before making decisions, and could for instance deter a vehicle from speeding up because to reduce the collective free energy, at the expense of reducing their own.

Extending CRE and VFE-based metrics hierarchically affords a natural and mathematically straightforward approach to first-principles AI safety. Several contemporary AI safety proposals feature prolific construction of probabilistic models (themselves constructed from AIs, at least in part). 
``Guaranteed Safe AI'' demands rigorous world modeling to construct formal safety guarantees \cite{dalrympleGuaranteedSafeAI2024, tegmarkProvablySafeSystems2023}. 
Bayesian, ``Scientist AIs'' exert caution within uncertainty bounds according to their world models, aided by simulation, but are also expected to require potentially massive amounts of compute \cite{bengioCautiousScientistAI, bengioCanBayesianOracle2024}. 
Elsewhere, the Gaia Protocol\footnotemark of globally coordinated, amortized learning, depends on LLM-aided context-dependent model construction \cite{leventovGaiaNetworkPractical2023, leventovGaiaNetworkIllustrated2024}.
There is strong overlap in each of these pursuits, grounded in the creation and exploration of probabilistic world models, and the VFE framework outlined herein provides a natural language to 1) embed safety specifications into world models, 2) direct agentic learning and exploration in their accordance, while 3) taking actions that are in the collective interest through the minimization of the joint free energy.

\footnotetext{Of which some of the authors are affiliated.}

\section{Gatekeeping Experiment}

We investigated the application of this principle in a simulated autonomous vehicle (AV) setting, using a pared-back simulator, \texttt{highway-env} \cite{highway-env}, which is built on top of \texttt{gymnasium} \cite{towers2024gymnasium}. Code for this experiment is available on \href{https://github.com/m-walters/av-agents}{Github} \cite{waltersAV2024}, and a sample video can be found \href{https://m-walters.github.io/assets/videos/double-anim.mp4}{here}.

Our highway track featured autonomous vehicles with a variable number of these being gatekeeper controlled. We adopt (and abuse) terminology from theory-of-mind research to distinguish Alters and Egos as the two main types of vehicles on the road. Alters have a static policy and constitute the background traffic of our simulation, whereas Egos are the vehicles of interest that we optionally assign gatekeepers to, measure, etc. Our results find that the introduction of gatekeepers controlling Ego policies according to CRE has an increasingly positive impact on the road as defined by stakeholder-defined preferences.

Each investigated configuration was seeded across 1200 world simulations, for a duration of 80 steps, which was enough time to allow traffic behaviors and consequences to emerge. 
When computing energy and risk estimates, every 5 world steps gatekeepers ran $N_{MC}=128$ internal Monte Carlo (MC) trajectories out to a $\tau=10$ step horizon. 
$N_{MC}$ is not exceedingly large, but for a relatively close horizon is sufficient for collecting an expectation of the upcoming future. 
The gatekeeper internal trajectories were fully observable---though their measurements naturally only considered neighbors within a reasonable radius.

\subsection{Rewards and Loss}
Our reward score was constituted from three aspects: target speed, collisions, and defensive driving.
Ego vehicles received a \textbf{speed reward} $R_S$ in the form of a Gaussian centered on a target speed $v_T$:
\begin{equation}
    \label{eq:R_S}
    R_S(v) = \alpha\exp[-(v-v_T)^2/2\sigma^2],
\end{equation}

where constants $\alpha$, $\sigma$, and $v_T$ were heuristically chosen. 
The \textbf{collision reward} $R_C$ was simply a constant based on collision state $s=s_c$,
\begin{equation*}
    R_C(s) = 
    \begin{cases} 
        -\kappa & \text{if } s=s_c \\
        0 & \text{otherwise}
    \end{cases}
\end{equation*}

with $\kappa$ heuristically chosen appropriately to ascribe high disincentive. 

Braking distance---the distance it takes to come to a full stop---is a property that scales quadratically with speed \cite{treiberCongestedTrafficStates2000}. This is an important property to capture, which we combine with the common sense that proximity is inherently more risky, to formulate our \textbf{defensive-driving reward}:
\begin{equation}
    \label{eq:R_D}
    R_D(j) = R_{D,max} - \lambda \sum_{i\in V_j}\frac{1}{2^m d_{ij}}\bigg[ w(i,j)^2 + \zeta \bigg], \\
\end{equation}
\begin{align*}
    w(i,j) = &\max(0, v_j-v_i)\times \mathrm{H}(x_i-x_j) \notag \\
    &+ \max(0, v_i - v_j)\times \mathrm{H}(x_j-x_i),
\end{align*}

with scalar $\lambda > 0$, vehicle index $i$ of vehicle $j$'s neighbors (set $V_j$), lane differential $m\in \{0, 1, 2, \dots\}$, and neighbor distance $d_{ij}$. $w(i,j)$ returns the magnitude of relative speed between $j$ and its neighbor, using the Heaviside binary function $\mathrm{H}$ to control for if a neighbor is in front or behind. If vehicles $j$ and $i$ are drifting apart, $w(i,j)$ is 0. The constant $\zeta$ adds an additional penalty for vehicle proximity. Since the terms are penalizing, we subtract the bulk from a max reward $R_{D,max}$ and truncate to the range $R_D(j)\in[0,R_{D,max}]$. The final result is a function that 1) penalizes quadratically with relative speeds between neighbors, 2) penalizes with increased proximity $\propto 1/d_{ij}$, but 3) less so as lane differential increases.

The fact that $R_C$ is negative is appropriately handled in the reward normalization process. Loss was then simply the negative sum of rewards, and constituted our only observed variable,
\begin{equation*}
    \calL = -\sum R_i.
\end{equation*}

It is worth highlighting here that the resulting improved road safety, as a consequence of gatekeeper decision-making, was achieved with this single aggregate scalar variable and did not require the suite of AV sensor inputs in its decision evaluation. 

\subsection{Risk formulation}
Since our experiment was a fully observable environment, and we assert \textit{ex hypothesi} that our loss and $\tilde{p}$ formulations are sufficient and accurate, we can drop any entropic contributions. 
In this context, therefore, CRE is identical to time-discounted expected utility\footnotemark.
Additionally, whereas the extrinsic terms in EFE/FEF are expectations over the variational model $q(x,o|\pi)$, we can directly work with $p(o,x|\pi)$ since we have a fully observable environment, and use Monte Carlo methods to approximate $p(o, x|\pi)$.
\footnotetext{
In future experiments involving partially-observable environments and other sources of uncertainty, the value of the complete CRE formulation given in \Eq{eq:risk-sum} will become more evident. See, for instance, \cite{kaufmannActiveInferenceModel2021, dacostaActiveInferenceDiscrete2020, parrActiveInferenceFree2022, tschantzLearningActionorientedModels2020, sajidActiveInferenceDemystified2021, ueltzhofferDeepActiveInference2018a, lanillosActiveInferenceRobotics2021, baltieriModularityActionPerception2018, baltieriActiveInferenceComputational2019, brownActiveInferenceAttention2011}.
}

The removal of entropy simplifies the determination of our stakeholder tolerance parameter. Without exploratory requirements, the scale of $beta$ is irrelevant---as with energy in many other contexts, we are only concerned with relative values, not absolutes\footnotemark.
In other applications, $\beta$ may be determined as a forced constraint: cost in dollars, quantity, etc.
\footnotetext{
The scale and shape of $\calG$ does become relevant when its absolute value matters, such as determining a risk threshold $\rho^*$.
}

Taken together, our final CRE is the expected utility
\begin{align}
    \calG_\Sigma(\calL) &= -\sum_{t'}^{\tau} \gamma^{t'} \E_{p(\calL)}[\ln\pref(\calL)] \notag \\
    &= \sum_{t'}^{\tau} \gamma^{t'} \E_{p(\calL)}[\calL] 
    \label{eq:risk}
\end{align}

\subsection{Policies}
The \texttt{highway-env} library has an automated \textit{Intelligent Driving Model} (\texttt{IDMVehicle}) \cite{treiberCongestedTrafficStates2000} class, which employs a combination of deterministic logic to calculate acceleration and steering. Lane changes are determined in part according to the \textit{Minimizing overall braking induced by lane change} ``MOBIL'' model \cite{kestingGeneralLaneChangingModel2007}, which, as advertised, tries to reduce imposed braking in its lane-change selection.

This vehicle policy is deterministic and has no machine learning or sampling involved in its decision-making. However, there are several knobs we can tune to produce different behavior types. For the Alter vehicles, we increased their appetite and aggression for lane changes, increasing the course difficulty for Egos. We also constructed two policies for Ego vehicles called ``Defensive'' and ``Hotshot''. These differ in their comfort with braking distance and lane-change aggression---Hotshot vehicles are more comfortable with tailgating a driver in front of them if it allows them to approach their target speed or get closer to a lane change they want. They are also more likely to accept an aggressive lane change.

In our experiment, a total of 24 vehicles were divided evenly between ego and alter vehicles. Among the ego vehicles, we experimented with different fractions of them being under GK control, also termed ``online''. In one configuration, 4/12 ego vehicles were online, and in another 12/12. Ego vehicles would start in the Hotshot policy, so in the 4/12 arrangement, the other 8 remained Hotshot for the entire run. Those under GK control were available for policy modulation between Hotshot and Defensive, based on the gatekeeper's CRE computation from simulated futures---like a driving instructor copilot that takes over when they anticipate upcoming danger. 

Since collisions are a metric of interest, conditions were set up such that these were not exceedingly rare events. During a run, 4 online ego vehicles would be tracked for a collision, the event of which would terminate a run. Additionally, if any 6+ vehicles were ever in a collision state, this would be considered a jam and the run terminated. Runs were not terminated on \textit{any} collision because it is still valuable to measure performance of ego vehicles in adapting to such road conditions. 

\subsection{Gatekeeper Policy Control}
\label{sec:gk-policy}
For online vehicles, gatekeepers anticipate upcoming risk through internal simulations, then toggle their vehicle's policy to Defensive in risky situations, or back to Hotshot when deemed safe enough. Using Hotshot as a nominal policy may seem odd, but it gives a stronger counterbalance to observe the phenomenon of interest\footnotemark. 
\footnotetext{As our construction could also apply to gatekeeping human drivers, the Hotshot policy is not a bad model of standard driver behavior in many parts of the world.}

Gatekeepers run $N_{MC}$ internal Monte Carlo trajectories at regular, frequent intervals in the world simulation to compute a CRE estimate, following \Eq{eq:risk}. Values for a given trajectory's risk are accumulated out to an MC horizon $\tau=10$ steps. 
Each vehicle's actions are not in a vacuum. Sharing local observations and predictions by opening channels of communication through gatekeepers enhances decision-making through a collective intelligence.
After computing individual CREs, we replace each with the average of their local neighborhoods and have online vehicles make policy decisions from this average. 

Converting from a unitless CRE value to a policy decision is not self-evident, and is open to the needs of the stakeholder. 
We opted for a simple CRE threshold method, where crossing the risk threshold, $\rho^*$, triggers a policy switch response.
To avoid erratic behavior at the threshold, i.e. frequent policy switching, we employed two thresholds $\rho^*_{+}=1.1\times \rho^*$ and $\rho^*_{-}=0.9\times \rho^*$, such that when risk crosses $\rho^*_{+}$ from below, the GK engages Defensive driving, and subsequently crossing $\rho^*_{-}$ from above engages Hotshot. Additionally, we used a 10-step graduation for policy parameter deltas, to smooth policy transitions and further reduce erratic behavior.
We selected a heuristic value of $\rho^*=2$, however determining $\rho^*$ is likely to be more straightforward in practical applications where the loss or CRE have units with bearing. 

\subsection{Results \& Discussion}
\begin{figure*}[!t]
    \centering
    \includegraphics[width=\textwidth]{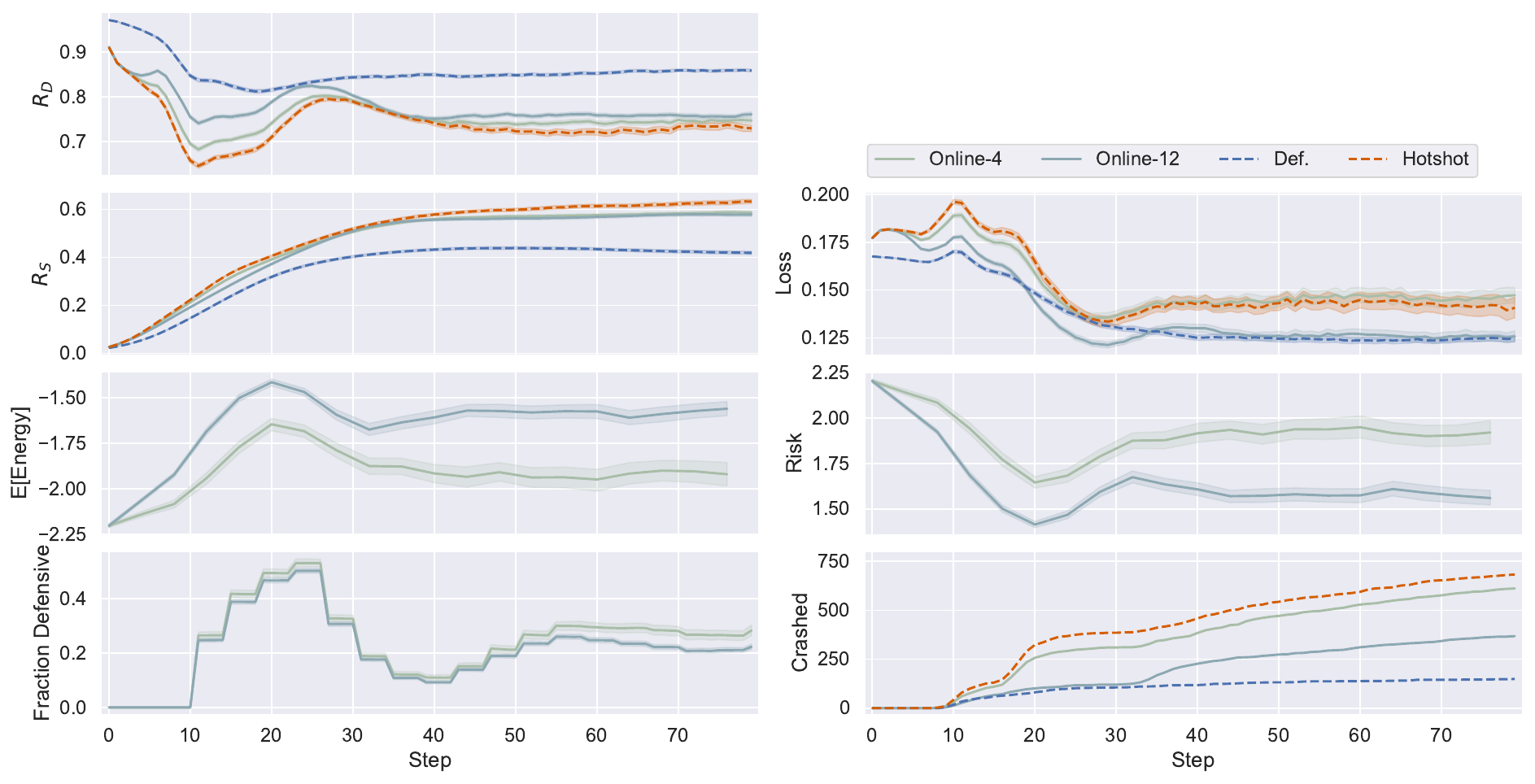}
    \caption{Baseline and gatekeeper results. Gatekeeper runs had either 4/12 or 12/12 ego vehicles online. $R_D$, $R_S$, Loss, Crashed, and Fraction Defensive are averaged realized values. Each $E$[Energy] and Risk measurement is across $N_{MC}$ MC trajectories. The Fraction Defensive is the proportion of ego vehicles in the Defensive policy. Crashed is a cumulation of how many worlds have had an ego crash at or before a given step. Values are averaged across 1200 world draws, 90\% CI displayed.}
    \label{fig:combined}
\end{figure*}
The ultimate goal here is better decision-making according to stakeholder preferences through simulated futures. To that end, our main measuring stick is the defined loss $\calL$ and collision results. 
Two baselines were simulated across 1200 world runs, for the Defensive and Hotshot policies. In a given baseline, all 12 of the ego vehicles would stick to the defined policy throughout, and thus no CRE calculations were performed for GK operations. Realized rewards and loss values were still measured at each step, however. 

Though the Hotshot policy has a consistently higher speed reward, it suffers in the defensive reward compared to the Defensive policy, and incurs substantially more collisions (\Fig{fig:combined}). Ultimately, the erratic, dangerous Hotshot behavior garners greater loss on average.

With the introduction of online gatekeepers, we aspire for the best of both worlds: \textit{intelligent policy selection that responds to environment conditions.} 
We found a considerable signal in support of this, that became increasingly pronounced proportional to GK presence. At full GK strength, crash avoidance was significantly improved, while finding opportunity to excel in defensive driving and target speed.

For the most part, the Defensive Baseline is always going to be hard to surpass: It is expected to have the fewest crashes and the highest $R_D$. 
Thus, gatekeepers need to perform comparatively well in those two dimensions while eking out gains in $R_S$---which is at odds with $R_D$ and $R_C$. Nonetheless, the Online-12 configuration handled this remarkably well, especially for the first half of the simulation where it tracked Hotshot-level $R_S$ while approaching Defensive-level $R_D$. 
This superior performance combination was most strongly exhibited in the Loss minimum by Online-12 around Step 25 that substantially outperformed both baselines.
From visual observations, the first half of the simulation is the more dynamic portion of the simulation, requiring egos to navigate around themselves and alters more (since they have a higher target speed than alters), versus the latter portion where the road approaches more of a steady-state.
(\href{https://m-walters.github.io/assets/videos/double-anim.mp4}{Example video}.)
The selection of $\rho^*=2$ yielded modest policy switching activity, and the ``Defensive Fraction'' in \Fig{fig:combined} indicates that typically between 10-40\% of egos would be in Defensive mode for the bulk of the run. 

Collisions (``Crashed'', \Fig{fig:combined}) could not be wholly eliminated, but these were present even in the Defensive baseline, so this is expected. The Online-4 configuration was slightly but measurably better than the Hotshot baseline in this, though Online-12 kept in tow with the Defensive baseline for the first half of the duration before diverging. In practical applications, if stakeholders want to push something like collision likelihood down even further, they need only update their preference prior, or the loss function penalty for collision, $\kappa$.

The Energy and Risk figures are from gatekeeper MC estimates. Risk calculations consider trajectories out to $\tau=10$ steps, so we should expect that early on with vehicles in Hotshot policy that it anticipates risk that reflects the baseline 10 steps ahead. Indeed this behavior tracks as the initial peak, subsequent dip, and plateau are anticipated by the Risk $\tau$ steps in advance. Trying to correlate spikes in Risk for Online modes with spikes in future baseline Loss becomes less accurate further into the simulation as their worlds continue to increasingly diverge after $t=0$.

The results show a clear trend: the effect of gatekeepers produces increasingly safer roads for everyone through superior driving according to our embedded preference. They can score highly in $R_S$, while incorporating smarter, safer driving when needed, reducing collisions, improving their $R_D$ scoring, and ultimately achieving better loss results than either baseline policy.

\section{Conclusion}

The Free Energy Principle, as one of the foundational underpinnings of Active Inference, draws powerful connections between physical energetic laws and intelligent action, with explanations for exploitation-exploration naturally emergent.
Encoding stakeholder preferences via the preference prior provides a highly flexible means to direct agentic learning. 
The Cumulative Risk Exposure metric introduced leverages these foundations to create an interpretable, modular utility to score policies according to biased futures.
The preference-temperature and tolerance mechanics outlined also introduce a conceptual and instructional foothold for usage. 

Stakeholders and AI agents can employ this safety metric to anticipate upcoming high risk situations and respond intelligently, as demonstrated by our autonomous vehicle experiment, which saw increasingly superior driving performance proportional to online usage.
This principle has immense potential across agentic applications as a quick and effective utility for gauging risk which, in contrast to simple loss measures, is biased towards stakeholder preferences, providing straightforward and transparent decision rules for risk governance and mitigation.

\printbibliography

@misc{
waltersAV2024,
    title = {Autonomous Vehicle Study Repository},
    author = {Walters, Michael},
    date = {2024-05-17},
    howpublished = {\url{https://github.com/m-walters/av-agents}},
}

@misc{highway-env,
  author = {Leurent, Edouard},
  title = {An Environment for Autonomous Driving Decision-Making},
  year = {2018},
  publisher = {GitHub},
  journal = {GitHub repository},
  howpublished = {\url{https://github.com/eleurent/highway-env}},
}

@misc{towers2024gymnasium,
      title={Gymnasium: A Standard Interface for Reinforcement Learning Environments}, 
      author={Mark Towers and Ariel Kwiatkowski and Jordan Terry and John U. Balis and Gianluca De Cola and Tristan Deleu and Manuel Goulão and Andreas Kallinteris and Markus Krimmel and Arjun KG and Rodrigo Perez-Vicente and Andrea Pierré and Sander Schulhoff and Jun Jet Tai and Hannah Tan and Omar G. Younis},
      year={2024},
      eprint={2407.17032},
      archivePrefix={arXiv},
      primaryClass={cs.LG},
      howpublished={\url{https://arxiv.org/abs/2407.17032}}, 
}

@article{bayesianBrain2010,
  author={Deutsch, Sid},
  journal={IEEE Pulse}, 
  title={Bayesian Brain: Probabilistic Approaches to Neural Coding (Doya, K., Eds., et al.; 2007) [Book Review]}, 
  year={2010},
  volume={1},
  number={3},
  pages={64-65},
  doi={10.1109/MPUL.2010.939182}
}

@misc{bengioCautiousScientistAI,
  author = {Bengio, Yoshua},
  title = {Towards a {{Cautious Scientist AI}} with {{Convergent Safety Bounds}}},  howpublished = {\url{https://yoshuabengio.org/2024/02/26/towards-a-cautious-scientist-ai-with-convergent-safety-bounds/}},
  langid = {english},
  date = {2024-02},
}

@inproceedings{baltieriActiveInferenceComputational2019,
  title = {Active {{Inference}}: {{Computational Models}} of {{Motor Control}} without {{Efference Copy}}},
  shorttitle = {Active {{Inference}}},
  booktitle = {2019 {{Conference}} on {{Cognitive Computational Neuroscience}}},
  author = {Baltieri, Manuel and Buckley, Christopher L.},
  date = {2019},
  publisher = {Cognitive Computational Neuroscience},
  location = {Berlin, Germany},
  doi = {10.32470/CCN.2019.1144-0},
  langid = {english}
}

@inproceedings{baltieriModularityActionPerception2018,
  title = {The Modularity of Action and Perception Revisited Using Control Theory and Active Inference},
  booktitle = {The 2018 {{Conference}} on {{Artificial Life}}},
  author = {Baltieri, Manuel and Buckley, Christopher L.},
  date = {2018},
  pages = {121--128},
  publisher = {MIT Press},
  location = {Tokyo, Japan},
  doi = {10.1162/isal_a_00031},
  eventtitle = {The 2018 {{Conference}} on {{Artificial Life}}},
  langid = {english}
}

@online{bengioCanBayesianOracle2024,
  title = {Can a {{Bayesian Oracle Prevent Harm}} from an {{Agent}}?},
  author = {Bengio, Yoshua and Cohen, Michael K. and Malkin, Nikolay and MacDermott, Matt and Fornasiere, Damiano and Greiner, Pietro and Kaddar, Younesse},
  date = {2024-08-22},
  eprint = {2408.05284},
  eprinttype = {arXiv},
  eprintclass = {cs},
  doi = {10.48550/arXiv.2408.05284},
  langid = {english}
}

@article{brownActiveInferenceAttention2011,
  title = {Active {{Inference}}, {{Attention}}, and {{Motor Preparation}}},
  author = {Brown, Harriet and Friston, Karl and Bestmann, Sven},
  date = {2011},
  journaltitle = {Frontiers in Psychology},
  shortjournal = {Front. Psychology},
  volume = {2},
  issn = {1664-1078},
  doi = {10.3389/fpsyg.2011.00218},
  langid = {english}
}

@article{dacostaActiveInferenceDiscrete2020,
  title = {Active Inference on Discrete State-Spaces: {{A}} Synthesis},
  shorttitle = {Active Inference on Discrete State-Spaces},
  author = {Da Costa, Lancelot and Parr, Thomas and Sajid, Noor and Veselic, Sebastijan and Neacsu, Victorita and Friston, Karl},
  date = {2020-12-01},
  journaltitle = {Journal of Mathematical Psychology},
  shortjournal = {Journal of Mathematical Psychology},
  volume = {99},
  pages = {102447},
  issn = {0022-2496},
  doi = {10.1016/j.jmp.2020.102447}
}

@online{dalrympleGuaranteedSafeAI2024,
  title = {Towards {{Guaranteed Safe AI}}: {{A Framework}} for {{Ensuring Robust}} and {{Reliable AI Systems}}},
  shorttitle = {Towards {{Guaranteed Safe AI}}},
  author = {Dalrymple, David and Skalse, Joar and Bengio, Yoshua and Russell, Stuart and Tegmark, Max and Seshia, Sanjit and Omohundro, Steve and Szegedy, Christian and Goldhaber, Ben and Ammann, Nora and Abate, Alessandro and Halpern, Joe and Barrett, Clark and Zhao, Ding and Zhi-Xuan, Tan and Wing, Jeannette and Tenenbaum, Joshua},
  date = {2024-07-08},
  eprint = {2405.06624},
  eprinttype = {arXiv},
  eprintclass = {cs},
  doi = {10.48550/arXiv.2405.06624}
}

@article{dayanHelmholtzMachine1995,
  title = {The {{Helmholtz Machine}}},
  author = {Dayan, Peter and Hinton, Geoffrey E. and Neal, Radford M. and Zemel, Richard S.},
  date = {1995-09-01},
  journaltitle = {Neural Computation},
  shortjournal = {Neural Computation},
  volume = {7},
  number = {5},
  pages = {889--904},
  issn = {0899-7667},
  doi = {10.1162/neco.1995.7.5.889}
}

@article{fristonDesigningEcosystemsIntelligence2024,
  title = {Designing {{Ecosystems}} of {{Intelligence}} from {{First Principles}}},
  author = {Friston, Karl J. and Ramstead, Maxwell J. D. and Kiefer, Alex B. and Tschantz, Alexander and Buckley, Christopher L. and Albarracin, Mahault and Pitliya, Riddhi J. and Heins, Conor and Klein, Brennan and Millidge, Beren and Sakthivadivel, Dalton A. R. and Smithe, Toby St Clere and Koudahl, Magnus and Tremblay, Safae Essafi and Petersen, Capm and Fung, Kaiser and Fox, Jason G. and Swanson, Steven and Mapes, Dan and Ren\'e, Gabriel},
  date = {2024-01},
  journaltitle = {Collective Intelligence},
  shortjournal = {Collective Intelligence},
  volume = {3},
  number = {1},
  eprint = {2212.01354},
  eprinttype = {arXiv},
  eprintclass = {nlin},
  issn = {2633-9137, 2633-9137},
  doi = {10.1177/26339137231222481}
}

@article{gottwaldTwoKindsFree2020,
  title = {The Two Kinds of Free Energy and the {{Bayesian}} Revolution},
  author = {Gottwald, Sebastian and Braun, Daniel A.},
  editor = {Gershman, Samuel J.},
  date = {2020-12-03},
  journaltitle = {PLOS Computational Biology},
  shortjournal = {PLoS Comput Biol},
  volume = {16},
  number = {12},
  issn = {1553-7358},
  doi = {10.1371/journal.pcbi.1008420},
  langid = {english}
}

@online{hadfield-menellCooperativeInverseReinforcement2024,
  title = {Cooperative {{Inverse Reinforcement Learning}}},
  author = {Hadfield-Menell, Dylan and Dragan, Anca and Abbeel, Pieter and Russell, Stuart},
  date = {2024-02-17},
  eprint = {1606.03137},
  eprinttype = {arXiv},
  eprintclass = {cs},
  doi = {10.48550/arXiv.1606.03137}
}

@inproceedings{
hyland2024freeenergy,
title={Free-Energy Equilibria: Toward a Theory of Interactions Between Boundedly-Rational Agents},
author={David Hyland and Tom{\'a}{\v{s}} Gaven{\v{c}}iak and Lancelot Da Costa and Conor Heins and Vojtech Kovarik and Julian Gutierrez and Michael J. Wooldridge and Jan Kulveit},
booktitle={ICML 2024 Workshop on Models of Human Feedback for AI Alignment},
year={2024},
url={https://openreview.net/forum?id=4Ft7DcrjdO}
}

@article{kaufmannActiveInferenceModel2021,
  title = {An {{Active Inference Model}} of {{Collective Intelligence}}},
  author = {Kaufmann, Rafael and Gupta, Pranav and Taylor, Jacob},
  date = {2021-07},
  journaltitle = {Entropy},
  volume = {23},
  number = {7},
  pages = {830},
  publisher = {Multidisciplinary Digital Publishing Institute},
  issn = {1099-4300},
  doi = {10.3390/e23070830},
  issue = {7},
  langid = {english}
}

@article{kestingGeneralLaneChangingModel2007,
  title = {General {{Lane-Changing Model MOBIL}} for {{Car-Following Models}}},
  author = {Kesting, Arne and Treiber, Martin and Helbing, Dirk},
  date = {2007-01-01},
  journaltitle = {Transportation Research Record},
  volume = {1999},
  number = {1},
  pages = {86--94},
  publisher = {SAGE Publications Inc},
  issn = {0361-1981},
  doi = {10.3141/1999-10}
}

@online{lanillosActiveInferenceRobotics2021,
  title = {Active {{Inference}} in {{Robotics}} and {{Artificial Agents}}: {{Survey}} and {{Challenges}}},
  shorttitle = {Active {{Inference}} in {{Robotics}} and {{Artificial Agents}}},
  author = {Lanillos, Pablo and Meo, Cristian and Pezzato, Corrado and Meera, Ajith Anil and Baioumy, Mohamed and Ohata, Wataru and Tschantz, Alexander and Millidge, Beren and Wisse, Martijn and Buckley, Christopher L. and Tani, Jun},
  date = {2021-12-03},
  eprint = {2112.01871},
  eprinttype = {arXiv},
  eprintclass = {cs},
  doi = {10.48550/arXiv.2112.01871}
}

@online{leibfriedVariationalInferenceModelFree2022a,
  title = {Variational {{Inference}} for {{Model-Free}} and {{Model-Based Reinforcement Learning}}},
  author = {Leibfried, Felix},
  date = {2022-12-18},
  eprint = {2209.01693},
  eprinttype = {arXiv},
  eprintclass = {cs},
  doi = {10.48550/arXiv.2209.01693}
}

@misc{leventovGaiaNetworkIllustrated2024,
  title = {Gaia {{Network}}: {{An Illustrated Primer}}},
  shorttitle = {Gaia {{Network}}},
  author = {Kaufmann, Rafael and Leventov, Roman},
  date = {2024-01-26},
  url = {https://forum.effectivealtruism.org/posts/BaoA3gz7xRaqn764J/gaia-network-an-illustrated-primer},
  langid = {english}
}

@misc{leventovGaiaNetworkPractical2023,
  title = {Gaia {{Network}}: A Practical, Incremental Pathway to {{Open Agency Architecture}}},
  shorttitle = {Gaia {{Network}}},
  author = {Kaufmann, Rafael and Leventov, Roman},
  date = {2023-12-20},
  url = {https://www.lesswrong.com/posts/AKBkDNeFLZxaMqjQG/gaia-network-a-practical-incremental-pathway-to-open-agency},
  langid = {english}
}

@article{mcmillenCollectiveIntelligenceUnifying2024,
  title = {Collective Intelligence: {{A}} Unifying Concept for Integrating Biology across Scales and Substrates},
  shorttitle = {Collective Intelligence},
  author = {McMillen, Patrick and Levin, Michael},
  date = {2024-03-28},
  journaltitle = {Communications Biology},
  shortjournal = {Commun Biol},
  volume = {7},
  number = {1},
  pages = {378},
  issn = {2399-3642},
  doi = {10.1038/s42003-024-06037-4},
  langid = {english}
}

@article{millidgeWhenceExpectedFree2021,
  title = {Whence the {{Expected Free Energy}}?},
  author = {Millidge, Beren and Tschantz, Alexander and Buckley, Christopher L.},
  date = {2021-02-01},
  journaltitle = {Neural Computation},
  shortjournal = {Neural Computation},
  volume = {33},
  number = {2},
  pages = {447--482},
  issn = {0899-7667},
  doi = {10.1162/neco_a_01354}
}

@book{parrActiveInferenceFree2022,
  title = {Active Inference: The Free Energy Principle in Mind, Brain, and Behavior},
  shorttitle = {Active Inference},
  author = {Parr, Thomas and Pezzulo, Giovanni and Friston, K. J.},
  date = {2022},
  publisher = {The MIT Press},
  location = {Cambridge, Massachusetts},
  isbn = {978-0-262-04535-3},
  langid = {english},
  pagetotal = {296}
}

@article{parrGeneralisedFreeEnergy2019,
  title = {Generalised Free Energy and Active Inference},
  author = {Parr, Thomas and Friston, Karl J.},
  date = {2019-12},
  journaltitle = {Biological Cybernetics},
  shortjournal = {Biol Cybern},
  volume = {113},
  number = {5--6},
  pages = {495--513},
  issn = {0340-1200, 1432-0770},
  doi = {10.1007/s00422-019-00805-w},
  langid = {english}
}

@online{richardsonLossInconsistencyProbabilistic2022,
  title = {Loss as the {{Inconsistency}} of a {{Probabilistic Dependency Graph}}: {{Choose Your Model}}, {{Not Your Loss Function}}},
  shorttitle = {Loss as the {{Inconsistency}} of a {{Probabilistic Dependency Graph}}},
  author = {Richardson, Oliver E.},
  date = {2022-02-24},
  eprint = {2202.11862},
  eprinttype = {arXiv},
  url = {http://arxiv.org/abs/2202.11862}
}

@article{sajidActiveInferenceDemystified2021,
  title = {Active Inference: Demystified and Compared},
  shorttitle = {Active Inference},
  author = {Sajid, Noor and Ball, Philip J. and Parr, Thomas and Friston, Karl J.},
  date = {2021-03},
  journaltitle = {Neural Computation},
  shortjournal = {Neural Computation},
  volume = {33},
  number = {3},
  eprint = {1909.10863},
  eprinttype = {arXiv},
  eprintclass = {cs},
  pages = {674--712},
  issn = {0899-7667, 1530-888X},
  doi = {10.1162/neco_a_01357}
}

@article{shreeshaStressSharingCognitive2024,
  title = {Stress Sharing as Cognitive Glue for Collective Intelligences: {{A}} Computational Model of Stress as a Coordinator for Morphogenesis},
  shorttitle = {Stress Sharing as Cognitive Glue for Collective Intelligences},
  author = {Shreesha, Lakshwin and Levin, Michael},
  date = {2024-10-30},
  journaltitle = {Biochemical and Biophysical Research Communications},
  shortjournal = {Biochemical and Biophysical Research Communications},
  volume = {731},
  pages = {150396},
  issn = {0006-291X},
  doi = {10.1016/j.bbrc.2024.150396}
}

@inproceedings{skalseLimitationsMarkovianRewards2023,
  title = {On the Limitations of {{Markovian}} Rewards to Express Multi-Objective, Risk-Sensitive, and Modal Tasks},
  booktitle = {Proceedings of the {{Thirty-Ninth Conference}} on {{Uncertainty}} in {{Artificial Intelligence}}},
  author = {Skalse, Joar and Abate, Alessandro},
  date = {2023-07-02},
  pages = {1974--1984},
  publisher = {PMLR},
  issn = {2640-3498},
  url = {https://proceedings.mlr.press/v216/skalse23a.html},
  eventtitle = {Uncertainty in {{Artificial Intelligence}}},
  langid = {english}
}

@online{tegmarkProvablySafeSystems2023,
  title = {Provably Safe Systems: The Only Path to Controllable {{AGI}}},
  shorttitle = {Provably Safe Systems},
  author = {Tegmark, Max and Omohundro, Steve},
  date = {2023-09-04},
  eprint = {2309.01933},
  eprinttype = {arXiv},
  eprintclass = {cs},
  doi = {10.48550/arXiv.2309.01933}
}

@online{treiberCongestedTrafficStates2000,
  title = {Congested {{Traffic States}} in {{Empirical Observations}} and {{Microscopic Simulations}}},
  author = {Treiber, Martin and Hennecke, Ansgar and Helbing, Dirk},
  date = {2000-08-30},
  eprint = {cond-mat/0002177},
  eprinttype = {arXiv},
  doi = {10.48550/arXiv.cond-mat/0002177}
}

@article{tschantzLearningActionorientedModels2020,
  title = {Learning Action-Oriented Models through Active Inference},
  author = {Tschantz, Alexander and Seth, Anil K. and Buckley, Christopher L.},
  editor = {Komarova, Natalia L.},
  date = {2020-04-23},
  journaltitle = {PLOS Computational Biology},
  shortjournal = {PLoS Comput Biol},
  volume = {16},
  number = {4},
  pages = {e1007805},
  issn = {1553-7358},
  doi = {10.1371/journal.pcbi.1007805},
  langid = {english}
}

@article{ueltzhofferDeepActiveInference2018a,
  title = {Deep Active Inference},
  author = {Ueltzh\"offer, Kai},
  date = {2018-12},
  journaltitle = {Biological Cybernetics},
  shortjournal = {Biol Cybern},
  volume = {112},
  number = {6},
  pages = {547--573},
  issn = {0340-1200, 1432-0770},
  doi = {10.1007/s00422-018-0785-7},
  langid = {english}
}

\end{document}